\documentclass[10pt,twocolumn,letterpaper]{article}

\usepackage[pagenumbers]{cvpr} 

\usepackage{makecell}    
\usepackage{pifont}  
\usepackage{booktabs} 
\usepackage{times}
\usepackage{epsfig}
\usepackage{graphicx}
\usepackage{amsmath}
\usepackage{siunitx}
\usepackage{xcolor}
\usepackage[utf8]{inputenc}
\usepackage{amssymb}
 \usepackage{multirow}
\usepackage[ruled,vlined]{algorithm2e}

\usepackage{times}  
\usepackage{helvet}  
\usepackage{courier}  
\usepackage[hyphens]{url}  
\usepackage{graphicx} 
\usepackage{natbib}  
\usepackage{caption} 

\usepackage[dvipsnames]{xcolor}

\makeatletter
\DeclareRobustCommand\onedot{\futurelet\@let@token\@onedot}
\def\@onedot{\ifx\@let@token.\else.\null\fi\xspace}

\makeatother

\definecolor{yellow}{rgb}{1,1, 0.6}
\definecolor{lightyellow}{rgb}{1,1, 0.8}
\definecolor{orange}{rgb}{1, 0.8, 0.6}
\definecolor{red}{rgb}{1, 0.6, 0.6}

\definecolor{wincolor}{rgb}{0.85, 0.0, 0.0}

\definecolor{darkyellow}{rgb}{0.8, 0.8, 0.5}
\definecolor{darkred}{rgb}{0.7, 0.3, 0.3}
\definecolor{darkgreen}{rgb}{0.3, 0.7, 0.3}
\definecolor{blue}{rgb}{0, 0, 1.0}
\definecolor{green}{rgb}{0, 1.0, 0}
\definecolor{pink}{rgb}{1, 0.4, 0.7}

\definecolor{customlightgray}{rgb}{0.95, 0.95, 0.95} %
\definecolor{darkgreen}{rgb}{0.0, 0.65, 0.0}
\definecolor{darkred}{rgb}{0.75, 0.0, 0.0} %
\definecolor{darkyellow}{rgb}{0.9, 0.72, 0} %
\definecolor{lightyellow}{rgb}{1, 1, 0.8}

\definecolor{DeltaColor}{rgb}{0.039,0.73,0.71}
\definecolor{SigmaColor}{rgb}{0.98,0.45,0.0}
\definecolor{AlphaColor}{rgb}{0,0,0.8}
\definecolor{BetaColor}{rgb}{0.8,0,0.8}
\definecolor{GammaColor}{rgb}{0.514,0.34,0.224}
\definecolor{EpsilonColor}{rgb}{0.353,0.725,0.906}
\definecolor{PurpleColor}{HTML}{9839ff}
\definecolor{RedColor}{rgb}{0.949,0.275, 0.224}
\definecolor{citecolor}{HTML}{0071bc}

\usepackage{pifont}
\usepackage{bbding}
\newcommand{\cmark}{{\ding{51}}\xspace}
\newcommand{\xmark}{{\ding{55}}\xspace}

\usepackage{xcolor}

\definecolor{cvprblue}{rgb}{0.21,0.49,0.74}
\usepackage[pagebackref,breaklinks,colorlinks,allcolors=cvprblue]{hyperref}

\def\paperID{3370} 
\def\confName{CVPR}
\def\confYear{2026}

\title{GeoMotion: Rethinking Motion Segmentation via Latent 4D Geometry}

\author{
Xiankang He$^{1,2}$\quad
Peile Lin$^{1,2}$\quad
Ying Cui$^{1,2}$\quad
Dongyan Guo$^{1,2}$\thanks{Corresponding author.}\\
Chunhua Shen$^{1,2,3}$\quad
Xiaoqin Zhang$^{1,2}$\\[4pt]
{\small $^{1}$ College of Computer Science and Technology, Zhejiang University of Technology}\\
{\small $^{2}$ Zhejiang Key Laboratory of Visual Information Intelligent Processing}\\
{\small $^{3}$ State Key Lab of CAD \& CG, Zhejiang University}
}

\begin{document}
\maketitle

\begin{abstract}
Motion segmentation in dynamic scenes is highly challenging, as conventional methods heavily rely on estimating camera poses and point correspondences from inherently noisy motion cues. Existing statistical inference or iterative optimization techniques that struggle to mitigate the cumulative errors in multi-stage pipelines often lead to limited performance or high computational cost. In contrast, we propose a fully learning-based approach that directly infers moving objects from latent feature representations via attention mechanisms, thus enabling end-to-end feed-forward motion segmentation. Our key insight is to bypass explicit correspondence estimation and instead let the model learn to implicitly disentangle object and camera motion. Supported by recent advances in 4D scene geometry reconstruction (e.g., $\pi^3$), the proposed method leverages reliable camera poses and rich spatial-temporal priors, which ensure stable training and robust inference for the model. Extensive experiments demonstrate that by eliminating complex pre-processing and iterative refinement, our approach achieves state-of-the-art motion segmentation performance with high efficiency. The code is available at:https://github.com/zjutcvg/GeoMotion.

\end{abstract}    
\section{Introduction}
\label{sec:intro}

Motion segmentation aims to disentangle moving objects from camera-induced motion in video sequences. As a fundamental problem in computer vision, it plays a vital role in autonomous driving, video understanding, robotics, 4D scene understanding, etc. However, the task remains highly challenging in dynamic real-world environments, where multiple objects, occlusions, and complex camera motion often coexist. While humans can effortlessly perceive moving objects even under occlusions or complex deformations, reproducing the ability in machines has proven difficult.

\begin{figure}[t]
    \centering  \includegraphics[width=0.475\textwidth]{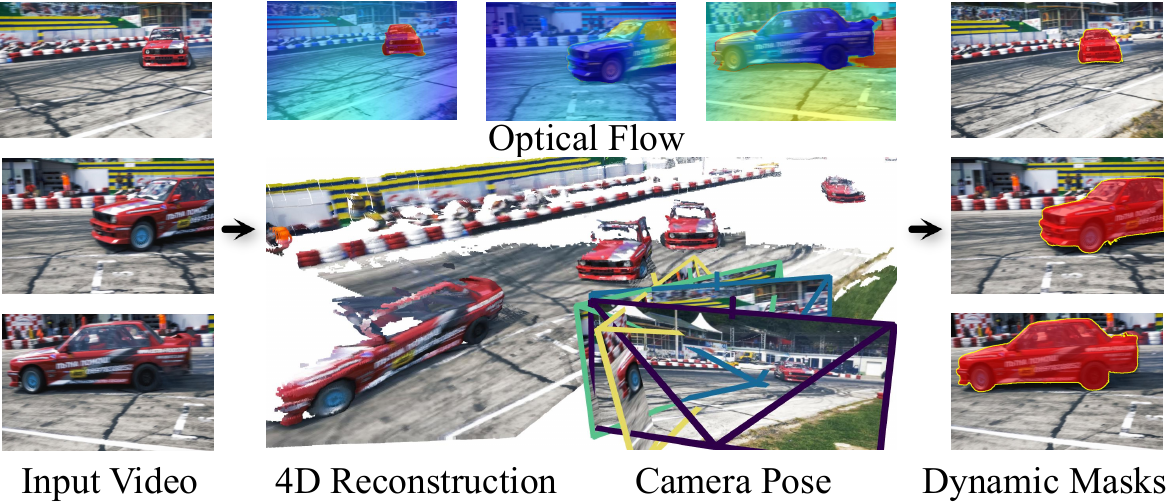}
    \caption{
    Overview of GeoMotion. Given an input video, our framework integrates 4D geometric priors from a pretrained reconstruction model ($\pi^3$) and local pixel-level motion from optical flow to infer dynamic object masks. By leveraging 4D geometric priors, the proposed GeoMotion disentangles object motion from camera motion in a single feed-forward manner.
    }
    \label{fig:teaser}
\end{figure}

Traditional models based on 2D motion cues such as optical flow~\cite{OCLR, xie24appearrefine}, struggle to distinguish independent object motion from camera motion due to the lack of depth difference. Moreover, the short temporal perception of optical flow make the models sensitive to occlusion and can not adapt to long-term requirements. To solve the issues, researchers devote to leveraging 3D depth information and iterative optimization to achieve robustness under unconstrained camera motion. RoMo~\cite{romo} explicitly combines optical flow with epipolar cues to disentangle object motion from camera motion, and utilizes iterative per-scene optimization for pose and mask refinement. MegaSaM~\cite{megasam} adopts a deep visual SLAM framework to estimate camera parameters and depth from dynamic monocular videos. Huang et. al. ~\cite{seganymotion} propose to combine long-range trajectory motion cues with SAM2 ~\cite{sam2} that enables pixel-level mask densification via iterative prompting. Despite their success, it is important to note that these iterative-optimization based models suffer significant drawbacks: 1) Heavy reliance on noise-prone intermediate representations. The pre-computed cues such as optical flow, point correspondences, or epipolar constraints, are often unreliable in dynamic scenes and lead to error accumulation throughout their multi-stage pipelines. 2) Computationally expensive iterative optimization. To tackle error accumulation, iterative optimization techniques such as iterative pose refinement, mask optimization or point tracking, are involved and result in significant inference overhead and poor scalability for real-world applications.

Recently, feed-forward models have shown remarkable progress in various vision analysis tasks such as visual segmentation ~\cite{sam, sam2}, 3D reconstruction ~\cite{dust3r,wang2025vggt} and 4D reconstruction ~\cite{chen2025easi3r, wang2025pi3}. The success raises a natural question: Can we also solve motion segmentation in a purely feed-forward manner? Thus these fundamental scene analysis tasks can be unified in almost one single framework. The ability of human perception that excels at segment anything is known to stem from a strong understanding of 3D scene geometry and spatio-temporal relationships. We therefore investigate whether such reasoning can be instantiated in learning through such 4D geometry priors. The answer is affirmative and we demonstrate that, for the first time, a much simple and efficient feed-forward model can achieve performance comparable to or even better than iterative optimization-based methods.

In the literature, feed-forward reconstruction models capable of estimating robust 3D and 4D scene attributes have opened new pathways for motion understanding. For example, both VGGT~\cite{wang2025vggt} and DUSt3R~\cite{dust3r} enables robust estimation of 3D geometry and camera parameters for images. Easi3R~\cite{chen2025easi3r} exploits the attention maps in DUSt3R to disentangle dynamic from static content for 4D reconstruction and eliminates the need for network fine-tuning. $\pi^3$~\cite{wang2025pi3} predicts affine-invariant camera poses and scale-invariant pointmaps without the need for a global coordinate system. These works confirm that pre-trained reconstruction models embed useful camera and motion-aware geometric priors. 
Therefore, the challenge reduces to learning how to decode these representations effectively for motion segmentation.

Our framework is designed with two modules: a feature aggregation module and a motion decoder module. 
The feature aggregation module integrates feature modalities including the latent 4D features and camera pose from 4D geometry priors, and the pixel-level motion in the optical flow. It constructs a unified spatio-temporal feature representation for motion mask learning.
We employ the standard alternating attention module from VGGT \cite{wang2025vggt} and $\pi^3$ \cite{wang2025pi3} as our visual geometry backbone to achieve the latent 4D geometric features that encode rich geometric and camera pose information. The camera pose decoder in $\pi^3$ is leveraged to further model the camera pose information.
The motion decoder module is simply composed of 5 self-attention layers to directly perceive dynamic objects from the aggregated feature representation. 
The whole framework is simple, neat and effective. The design allows the model to achieve motion disentanglement within a single feed-forward pass. 
Figure~\ref{fig:teaser} illustrates a visualization of the model.
During the testing phase, we furthermore employ the visual segmentation SAM2 ~\cite{sam2} to improve the resolution of the predicted motion masks. The operation is the same as it is in the final refinement of RoMo \cite{romo}, but different from SegAnyMotion~\cite{seganymotion} that using SAM2 as iterative prompting.

In summary, the contributions of this work are as follows:
\begin{itemize}
\itemsep 0pt
	\item We present a feed-forward motion segmentation framework that directly leverages 4D geometric priors without iterative optimization. To our knowledge, it is the first efficient feed-forward model that achieves comparable performance with iterative-optimization based motion segmentation methods. 

    \item  By directly learning motion from 4D latent geometry, we demonstrate that noisy intermediate correspondence estimation can be eliminated, thus enabling accurate motion segmentation without iterative optimization.
 	
    \item Without bells and whistles, the proposed approach achieves state-of-the-art accuracy on multiple challenging benchmarks. Being substantially simpler and faster than existing optimization-based methods, it establishes a new paradigm for geometry-informed feed-forward motion understanding.
\end{itemize}

\section{Related Work}
\label{sec:Related}

\subsection{Conventional Motion Segmentation}

Motion segmentation aims to predict dynamic object masks from video inputs. Existing predominant approaches rely on estimating camera poses and point correspondences from estimated explicit motion cues, primarily through optical flow~\cite{meunier2022driven, RCF, yang2021self, xie2022segmenting} or point trajectories~\cite{brox2010object, karazija2025learning, ochs2013segmentation, sheikh2009background, yan2006general, point-2frame-1, point-2frame-2, point-2frame-3}.
While informative in constrained environments, such motion cues are prone to noise and instability under challenging conditions including texture-less surfaces, occlusions, and significant camera motion.
To address these issues, flow-based methods~\cite{yang2021self, xie2022segmenting, RCF, EM} often integrate occlusion reasoning or appearance information to improve flow coherence, while trajectory-based techniques~\cite{affinities-1} aggregate long-term correspondences over extended temporal windows to strengthen motion consistency.
However, both paradigms remain fundamentally limited by inaccuracies in their underlying motion estimation. Noisy flow fields and trajectory drift propagate through multi-stage architectures, resulting in cumulative errors and segmentation instability.
Recent efforts mitigate these effects through iterative optimization. For instance, RoMo~\cite{romo} builds on RAFT~\cite{RAFT} to achieve optical flows
and iteratively refines motion segmentation using epipolar constraints and segmentation priors from models like SAM and SAM2 \cite{sam, sam2}.
Similarly, SegAnyMotion~\cite{seganymotion} adopts point trajectories as prompts within an iterative segmentation loop.
Although these iterative approaches enhance robustness, they incur significant computational cost and hindering deployment in real-world scenarios.
In contrast to existing paradigms, we propose a feed-forward framework that bypasses explicit motion computation altogether. By leveraging latent 4D geometric representations, our approach implicitly disentangles object motion from camera motion without iterative refinement, enabling accurate and efficient segmentation under dynamic and complex scene conditions.

\subsection{Feed forward 4D Reconstruction}
4D scene reconstruction focuses on recovering temporally consistent 3D scene geometry and the corresponding camera pose from dynamic video inputs. It forms a critical foundation for holistic dynamic scene understanding. Substantial progress for it has been achieved in recent years~\cite{lu2024align3r,point3r,wang2024spann3r,das3r,Yao_2025_CVPR,jiang2025geo4d}, with methods broadly falling into two categories: optimization-based and feed-forward approaches.
Methods such as MonST3R~\cite{monst3r} and C4D~\cite{c4d} adopt an optimization-based paradigm to jointly optimize depth, motion, and camera pose through iterative refinement. While achieving notable accuracy, these methods suffer from high computational cost and limited scalability for real-world applications. 

Benefiting from large-scaled data resources, feed-forward reconstruction methods have emerged as an efficient and scalable alternative. DUSt3R~\cite{dust3r} introduces a transformer-based architecture for pairwise 3D reconstruction in static scenes. CUT3R~\cite{cut3r} further extends this paradigm with a stateful recurrent design, while TTT3R~\cite{ttt3r} explores test-time training strategies to enhance generalization across image streams. PAGE-4D~\cite{page4d} extends it to dynamic settings via a soft dynamics-aware mask that modulates attention for camera pose and 3D geometry predicting. However, its single-mask design limits fine-grained motion modeling. Easi3R~\cite{chen2025easi3r} devotes to extract motion cues from static models in a training-free manner, yet operates post-hoc and lacks end-to-end integration. VGGT4D~\cite{vggt4d} extends the Easi3R philosophy to the VGGT~\cite{wang2025vggt} backbone, and MUT3R~\cite{mut3r} further adapts similar ideas to CUT3R. Both approaches extract dynamic cues from foundation models that were originally designed for static reconstruction, rather than explicitly modeling dynamic scene structure. $D^2USt3R$~\cite{d2ust3r} enables dense, temporally aligned point prediction under dynamic supervision, yet operates on fixed-view pairs and lacks camera pose estimation. 

Among feed-forward approaches, $\pi^3$~\cite{wang2025pi3} stands out as a unified solution that jointly predicts per-frame camera poses and dense point clouds without relying on fixed reference views. Its use of permutation-equivariant attention and alternating view-wise and global self-attention yields temporally coherent geometry while encoding rich motion-aware features. 
The explicit camera motion and consistent 4D geometry modeling in $\pi^3$ offer strong priors for motion understanding. Our work leverages these rich 4D geometry priors to learn dynamic object masks through the integrated fusion of geometry, flow, and camera pose representations within a unified feed-forward framework.

\section{Method}

\begin{figure*}[t]
    \centering  \includegraphics[width=1.0\textwidth]{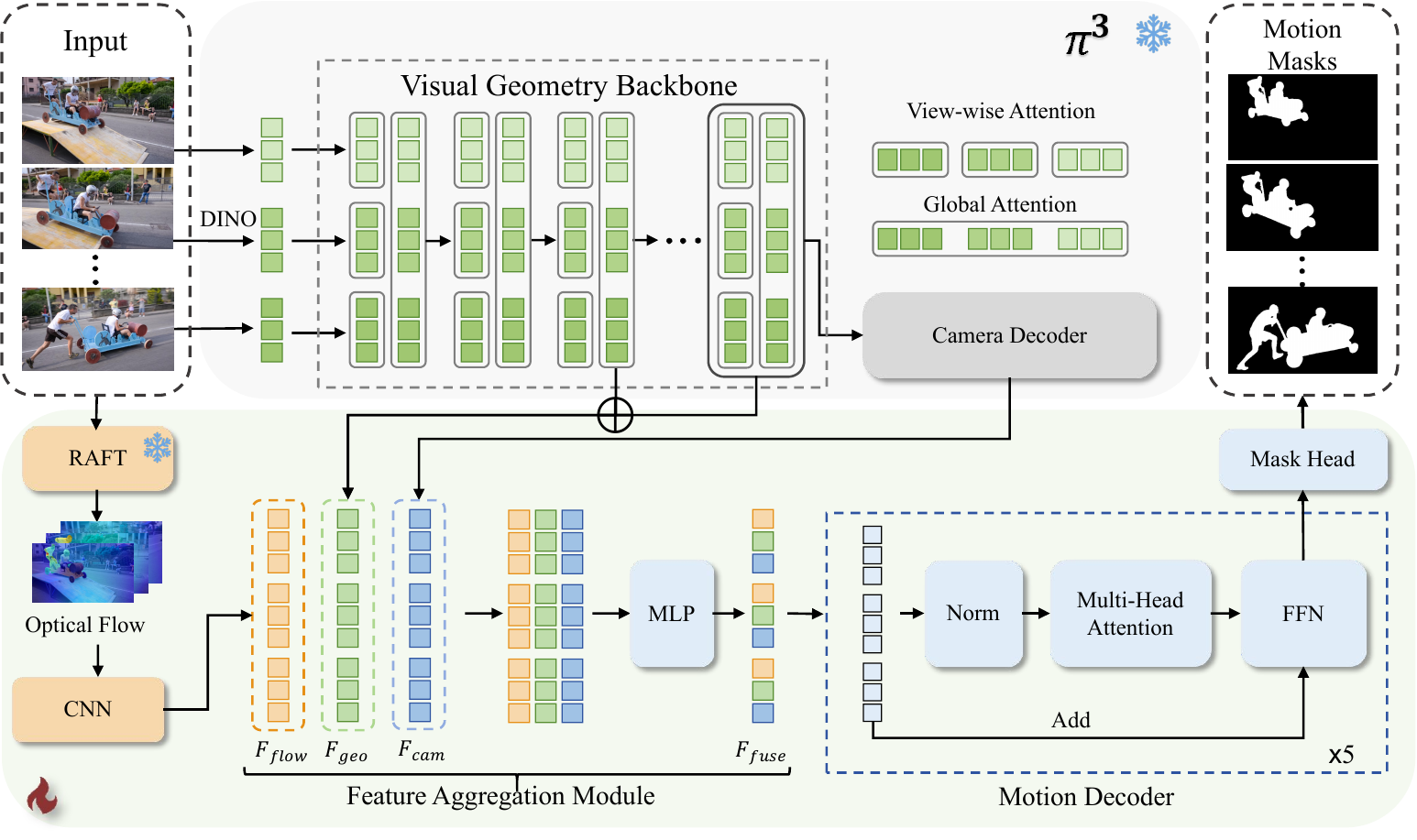}
    \caption{
    \textbf{Architecture of the proposed GeoMotion framework.}
     The model comprises a feature aggregation module and a motion decoder. The former fuses latent 4D features, optical flow features, and camera pose embeddings, while the latter employs multi-head self-attention to decode motion masks. The design enables end-to-end feed-forward motion segmentation without iterative refinement.
    }
    \vspace{-1em}
    \label{fig:method}
\end{figure*}

We tackle motion segmentation by fundamentally rethinking its formulation as directly learning motion from geometry representations, thus bypassing error accumulation from imperfect intermediate estimations and computationally expensive iterative optimization. A feed-forward framework, namely GeoMotion, is designed by leveraging rich 4D geometric priors encoded in the pre-trained 4D reconstruction model, and then decoding the motion masks by applying self-attention exclusively to the feature representations. 
In the following, we describe its components in detail.

\subsection{Feed-forward Architecture}

Figure~\ref{fig:method} illustrates the overall architecture of the proposed framework. It simply contains a feature aggregation module to acquire necessary video features, and a motion decoder module to directly learn motion masks from the features.

Optical flow provides pixel-level motion information, which can be considered as a composite signal of both camera pose variation and dynamic object motion. 
In parallel, the 4D scene reconstruction model like $\pi^3$,  which is trained on large-scale datasets,  acquires the capability to generate geometric output such as point clouds and camera pose predictions. As a result, the feature layers in its model inherently encode rich priors about scene structure, 3D geometry, and camera pose information, which can be considered as latent 4D geometry features. These 4D priors and pixel-level motion features in the optical flow are highly complementary to each other. 
By integrating these feature modalities, our model constructs a unified spatio-temporal feature representation that combines local pixel motion, global scene geometry, and camera dynamics, thus enabling directly decoding moving objects without explicit correspondence matching or iterative optimization.

Given a sequence of $N$ images, the features per-image are extracted by DINOv2 \cite{oquab2023dinov2}.
We then use the alternating attention module of $\pi^3$ as our visual geometry backbone (VGB) to achieve the latent 4D geometry features $\mathbf{F}_{\rm geo}$ from these per-image features. 
The camera pose decoder in $\pi^3$ is leveraged after the VGB to model the camera pose $\mathbf{F}_{\rm cam}$.
In parallel, we use RAFT \cite{RAFT} to obtain optical flows, which are transformed by CNN to perceive local optical flow features $\mathbf{F}_{\rm flow}$. 
Our feature aggregation module then fuses the three feature modalities through a simple MLP operation:
\begin{equation}
    \mathbf{F}_{\rm fuse} = \mathrm{MLP}([\mathbf{F}_{\rm geo}; \mathbf{F}_{\rm flow}; \mathbf{F}_{\rm cam}]).
\end{equation}
Details on the aggregation of three feature modalities are provided in the Supplementary Materials.

The motion decoder module is simply composed of 5 self-attention layers. It directly perceives dynamic objects from the fused feature representations. The motion masks are generated by a lightweight MLP head, which takes the output of the last layer in the decoder as input. 

In the architecture, the pre-trained weights of DINO and RAFT are loaded from their models. The parameters of the VGB and camera pose decoder are loaded from the pre-trained weights for the $\pi^3$ alternating-attention layers and camera pose decoder layers. They are all frozen during training. 
The parameters of the motion decoder module is initialized with the pre-trained weights of the $\pi^3$ confidence decoder.
The whole framework is simple, neat, and works in an end-to-end feed-forward manner. 

Since the feature representations are of low-resolution compared to the original frame, the motion masks are coarse ones. During the testing phase, the predicted motion masks are put into a visual segmentation model SAM2 \cite{sam2} to achieve the final fine-grained segmentation masks of high resolution. The operation is the same as it in the final refinement of RoMo \cite{romo}.

\subsection{Visual Geometry Backbone}

\begin{figure}[t]
    \centering  \includegraphics[width=0.47\textwidth]{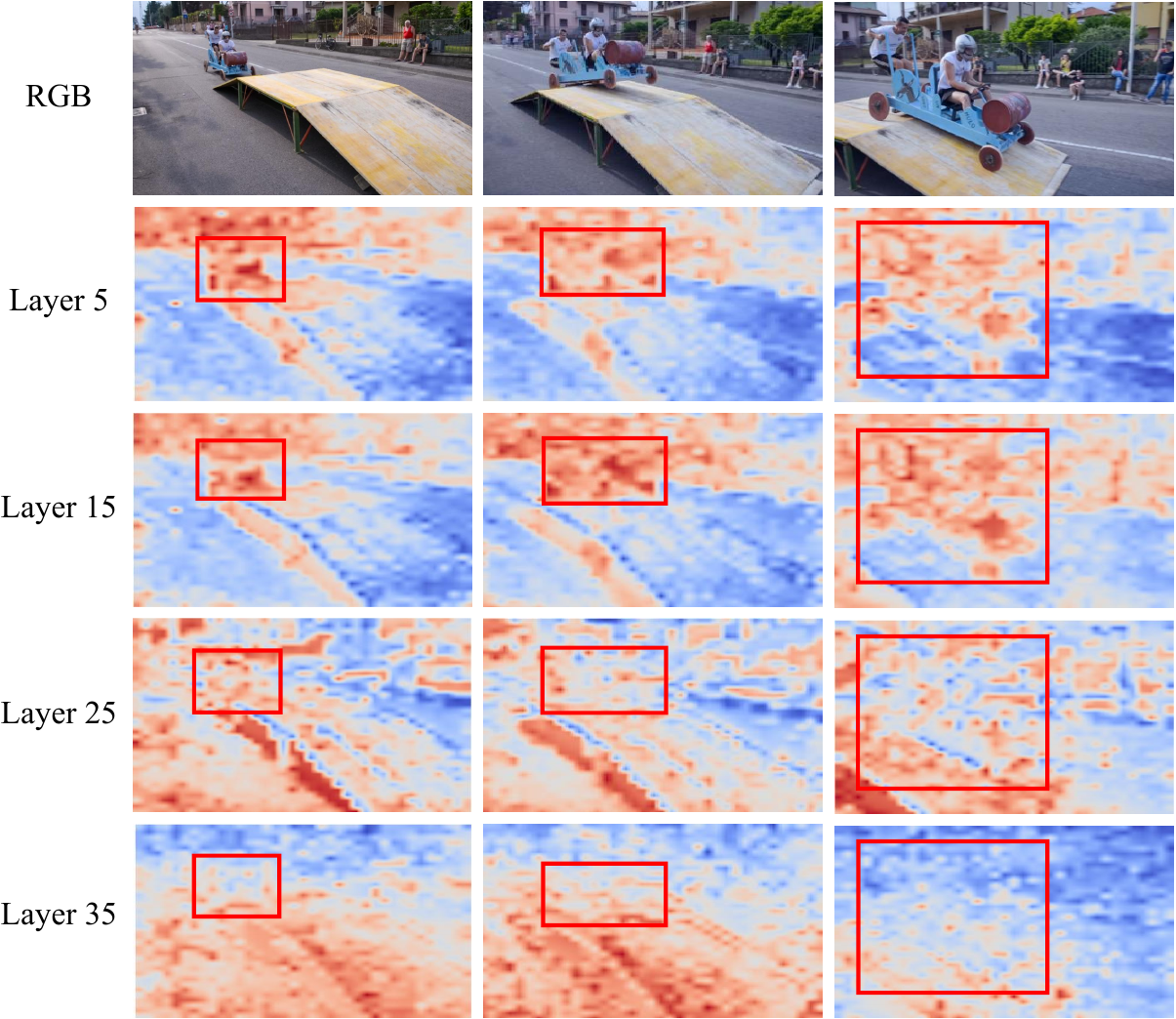}
    \caption{
    Visualization of $\pi^3$ features across alternating attention layers. Shallow layers preserve semantic object-level features, whereas deeper layers encode high-level global geometry. Their fusion yields robust latent 4D representations that support accurate motion segmentation.
    }
    \label{fig:layer}
\end{figure}

We employ the standard alternating attention module from VGGT \cite{wang2025vggt} and $\pi^3$ \cite{wang2025pi3} as our visual geometry backbone. The same as in $\pi^3$, we use 36 layers for the alternating attention to perceive latent 4D geometry features.
The shallow layers that are closer to the DINO features tend to perceive image-level features. The deeper layers accumulating high-level features layer by layer are more inclined to perceive richer contextual and structural patterns. A visualization of features in these layers is shown in Figure~\ref{fig:layer}.
As a consequence, we integrate both shallow and deeper layer features to achieve robust latent 4D features. The output feature $\mathbf{F}_{\rm geo}$ is formed with empirically selected $5^{th}, 15^{th}, 35^{th}$ and $36^{th}$ layers through concatenation. Moreover, the output of the last two layers in VGB is sent to the camera pose decoder to yield the camera pose estimation $\mathbf{F}_{\rm cam}$.

\subsection{Training Loss}

The mask decoder predicts a motion mask $M \in [0,1]^{H \times W}$ for each input video frame.
It represents the probability of each pixel belonging to a moving object. During training, the prediction is supervised by its corresponding ground-truth binary motion mask $M_{\text{gt}}$, where moving objects are labeled as $1$ and background pixels as $0$. The training objective $\mathcal{L}$ combines Focal Loss~\cite{focal} and Dice Loss~\cite{diceloss} over a sequence of $N$ frames:
\begin{equation}
\mathcal{L} = \sum_{t=1}^{N} \left( \lambda_1 \mathcal{L}_{\text{focal}}(M^{t}, M_{\text{gt}}^{t}) + \lambda_2 \mathcal{L}_{\text{dice}}(M^{t}, M_{\text{gt}}^{t}) \right)
\end{equation}
where $\lambda_1$ and $\lambda_2$ balance the contribution of each loss term. 
Focal Loss helps the model to emphasize hard-to-classify pixels, thereby improving its robustness to small, ambiguous, or partially occluded moving objects. 
Dice Loss mitigates the foreground-background class imbalance by optimizing the spatial overlap between predictions and ground-truth, thereby achieving more stable and coherent motion prediction.

Together with the multi-frame attention in the motion decoder and the 4D geometric priors provided by $\pi^3$, the composite loss enables robust motion segmentation that effectively distinguishes object motion from camera-induced parallax.

\begin{table*}[t]
\small
\centering
\caption{
Quantitative comparison with motion segmentation methods on popular benchmarks.
The proposed model obtains state-of-the-art performance. It achieves an excellent trade-off between segmentation quality and computational efficiency.
}
\vspace{-0.5em}
\setlength{\tabcolsep}{5pt}
\renewcommand{\arraystretch}{1}
\resizebox{\textwidth}{!}{
    \begin{tabular}{@{}lccccccccccccccc@{}}
    \toprule
    \multirow{2}{*}{Methods} 
    & \multirow{2}{*}{\begin{tabular}{@{}c@{}}Iterative\\optimization\end{tabular}} 
    & \multicolumn{3}{c}{DAVIS2016-M} 
    & \multicolumn{1}{c}{DAVIS2017} 
    & \multicolumn{2}{c}{FBMS-59}
    & \multicolumn{3}{c}{DAVIS2016}
    & \multicolumn{1}{c}{SegTrackV2}
    & \multirow{2}{*}{\begin{tabular}{@{}c@{}}Run Time\\per frame\end{tabular}} \\
    
    \cmidrule(l){3-5} \cmidrule(l){6-6} \cmidrule(l){7-8} \cmidrule(l){9-11} \cmidrule(l){12-12}
    &  
    & $\mathcal{J\&F} \uparrow$ & $\mathcal{J} \uparrow$ & $\mathcal{F}\uparrow$
    & $\mathcal{J} \uparrow$ 
    & $\mathcal{J} \uparrow$ & $\mathcal{F}\uparrow$
    & $\mathcal{J\&F} \uparrow$ & $\mathcal{J} \uparrow$ & $\mathcal{F}\uparrow$
    & $\mathcal{J} \uparrow$
    & \\ 
    \midrule
    OCLR-TTA~\cite{OCLR} & Yes & 78.5 & 80.2 & 76.9 & 76.0  & 69.9 & 68.3 & 78.8 & 80.8 & 76.8 & 72.3 & 1.25s \\
    RoMo~\cite{romo} & Yes & - & - & - & -  & 75.5 & -  & - & 77.3 & - & 67.7 & 8.34s \\
    SegAnyMotion~\cite{seganymotion} & Yes & 89.5 & 89.2 & 89.7 & 90.0  & 78.3 & 82.8  & 90.9 & 90.6 & 91.0 & 76.3 & 6.44s \\
    
    \midrule
    CIS~\cite{yang2019CIS} & No & 66.2 & 67.6 & 64.8 & -  & 63.6 & - & 68.6 & 70.3 & 66.8 & 62.0 & - \\
    EM~\cite{EM} & No & 75.2 & 76.2 & 74.3 & 55.5  & 57.9 & 56.0 & 70.0 & 69.3 & 70.7 & 55.5 & - \\
    RCF-Stage1~\cite{RCF} & No & 77.3 & 78.6 & 76.0 & 76.7  & 69.9 & - & 78.5 & 80.2 & 76.9 & 76.7 & - \\
    OCLR-flow~\cite{OCLR} & No & 70.0 & 70.0 & 70.0 & 69.9  & 65.5 & 64.9 & 71.2 & 72.0 & 70.4 & 67.6 & 0.20s \\
    ABR~\cite{xie24appearrefine} & No & 72.0 & 70.2 & 73.7 & 74.6  & \textbf{81.9} & \textbf{79.6} & 72.5 & 71.8 & 73.2 & 76.6 & 0.28s \\
    Ours & No & \textbf{83.9} & \textbf{83.5} & \textbf{84.3} & \textbf{81.1}  & 72.5 & 78.5 & \textbf{84.7} & \textbf{84.5} & \textbf{85.0} & \textbf{77.3} & 0.31s \\
    \bottomrule
    \end{tabular}
}
\label{tab:mos}
\vspace{-1em}
\end{table*}

\begin{figure*}[t]
    \centering
    \includegraphics[width=1.0\textwidth]{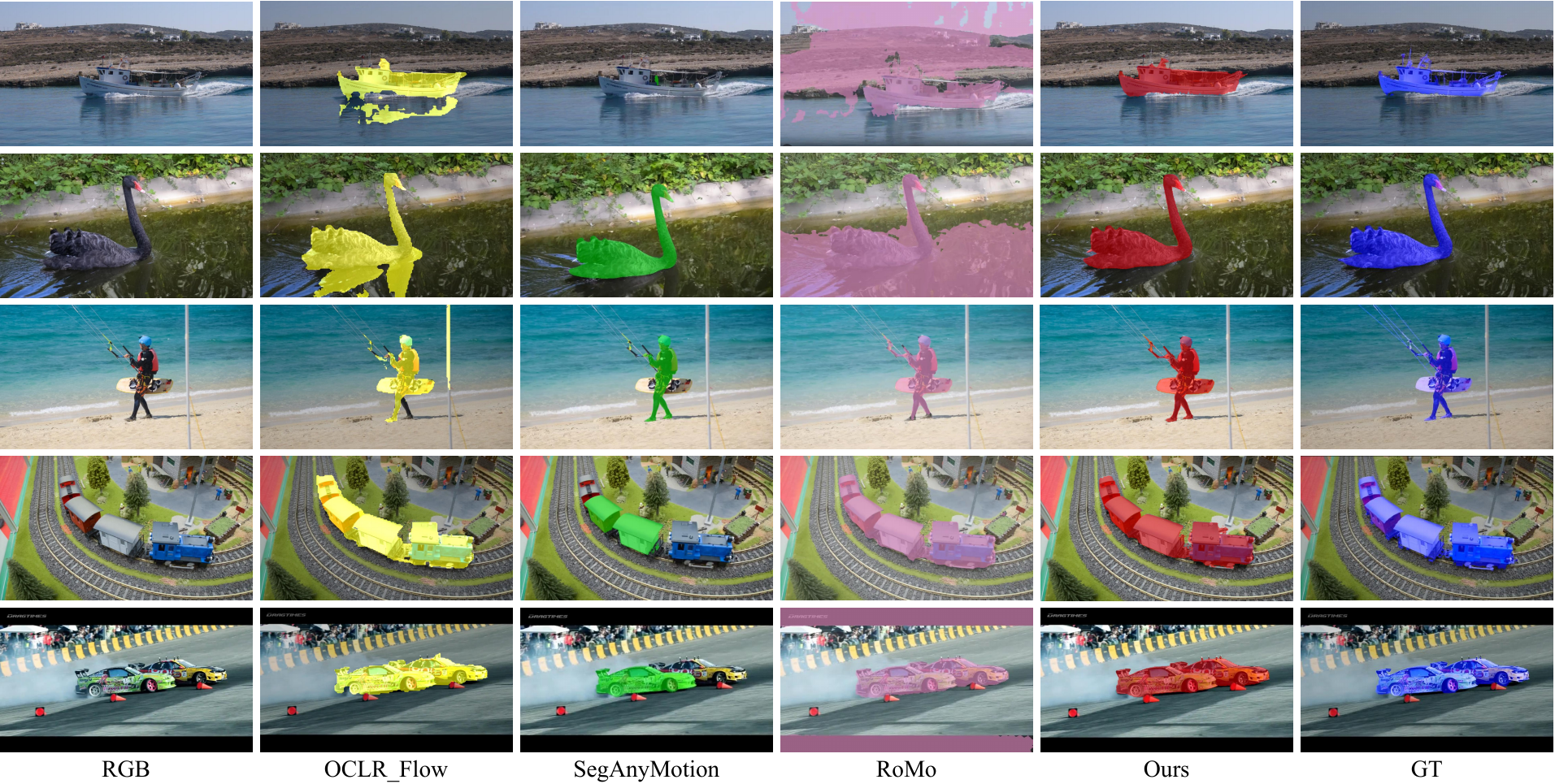}
     \vspace{-2em}
    \caption{\textbf{Qualitative comparison on multiple benchmarks.} 
    Visual comparison with state-of-the-art methods including OCLR-Flow~\cite{OCLR}, SegAnyMotion~\cite{seganymotion}, and RoMo~\cite{romo}. The proposed method produces geometrically complete and visually coherent motion masks, preserving fine object details and boundaries under complex scenes.
    }
    \label{fig:vis}
    \vspace{-1 em}
\end{figure*}

\section{Experiment}

\subsection{Experimental Settings}
\noindent \textbf{Training Datasets.} 
We collect training data from five public dynamic-scene datasets: HOI4D~\cite{hoi4d}, Dynamic Replica~\cite{DynamicReplica}, YouTubeVOS2018-motion~\cite{xie24appearrefine}, and selected videos from OmniWorld-Game~\cite{zhou2025omniworld} and GOT-10K~\cite{got10k}.
\textbf{HOI4D} is an egocentric real-world dataset with 2972 videos that captures diverse human–object interaction scenarios with official motion segmentation annotations. In our experiments, we utilize only frames containing dynamic object masks.
\textbf{Dynamic Replica} is a synthetic dataset designed for 3D reconstruction with both camera and object motion. It consists of 584 videos and offers long-term tracking annotations and object masks. We follow the pre-processing protocol of Seganymotion~\cite{seganymotion} to obtain dynamic masks.
YouTubeVOS2018-motion~\cite{xie24appearrefine} (short form \textbf{``YTVOS18-m"}) contains 120 sequences with one or more dominant moving objects. As a video object segmentation dataset, it provides rich annotated sequences and dynamic masks. We adopt the same pre-processing strategy from ABR~\cite{xie24appearrefine}.
We further formulate a \textbf{OmniWorld-motion} dataset and a \textbf{GOT-Motion} dataset. OmniWorld-motion contains 9 sequences with about $2000 \sim 10000$ frames per video. They are selected from the OmniWorld-game \cite{zhou2025omniworld} dataset, which contains diverse scenes with long temporal sequences. 
GOT-Motion is a subset of GOT-10K \cite{got10k}, which is a challenging real-world object tracking dataset. To form it, we select 130 short videos with average 97 frames per sequence, and annotate motion masks to enhance the diversity of dynamic scenes in our training set.
During training, we sample videos from all datasets and select different frames across epochs to ensure diverse scene coverage and motion variation.

\noindent \textbf{Benchmarks and Metrics.} 
For evaluation, we use five popular motion segmentation benchmarks: DAVIS2016~\cite{davis_2016}, DAVIS2016-Moving~\cite{dave2019towards}, DAVIS2017~\cite{davis_2017}, SegTrack-v2~\cite{segtrackv2}, and FBMS-59~\cite{fbms59}.
All evaluation datasets remain unseen during training, enabling a zero-shot assessment of the model’s generalization ability across diverse dynamic scenarios.
Following prior works~\cite{romo, seganymotion, xie24appearrefine}, we evaluate region similarity ($\mathcal{J}$), contour similarity ($\mathcal{F}$) and their average ($\mathcal{J\&F}$), region recall ($\mathcal{J_R}$), and mean IoU ($\mathcal{J_M}$).
Specifically, $\mathcal{J}$ and $\mathcal{F}$ measure spatial overlap and boundary accuracy, respectively, and their combination ($\mathcal{J\&F}$) provides a balanced assessment of segmentation quality~\cite{segtrackv2, davis_2017}.
$\mathcal{J_R}$ denotes the percentage of frames with $\mathcal{J} > 0.5$, reflecting the model’s ability to recall moving regions~\cite{chen2025easi3r}, while $\mathcal{J_M}$ represents the mean IoU, following conventions in reconstruction-based methods. These metrics together ensure a fair and comprehensive comparison across methods and datasets.

\noindent \textbf{Implementation.}
We train our model for 15 epochs using the Adam optimizer~\cite{adam} with a learning rate of $5e^{-5}$ on 4 NVIDIA RTX 5090 GPUs (32GB each). To capture diverse motion patterns, the frame sampling step is dynamically adjusted according to the average frames of videos in each dataset.
During training, each batch contains 16 frames. The frames are center-cropped to maintain the aspect ratio, and resized to 518$\times$518 resolution. To enhance data diversity, frames are randomly re-sampled for each video at each epoch, ensuring that different frames are used across training.
The balance parameters $\lambda_1$ and $\lambda_2$ are equally set as $0.5$.

\subsection{Comparison with Motion Segmentation Methods}
\label{sec:sota}

Our approach is compared with representative motion-cue-based methods, including CLS~\cite{yang2019CIS}, EM~\cite{EM}, RCF~\cite{RCF}, OCLR-flow~\cite{OCLR}, and ABR~\cite{xie24appearrefine}, as well as iterative optimization-based methods, including OCLR-TTA~\cite{OCLR}, RoMo~\cite{romo}, and SegAnyMotion~\cite{seganymotion}. 
In terms of supervision, OCLR-TTA and OCLR-flow are trained exclusively on synthetic data, with OCLR-TTA additionally performing test-time adaptation. RoMo, CIS, EM, and RCF are fully unsupervised and trained on unlabeled videos. ABR is pre-trained on synthetic data and further adapted to real videos in a self-supervised manner. SegAnyMotion relies on large-scale human-annotated motion segmentation data.

\subsubsection{Quantitative Comparison}
A quantitative comparison is shown in Table~\ref{tab:mos}. 
The results demonstrate that the proposed method achieves superior segmentation accuracy and robustness across multiple benchmarks while maintaining high efficiency.
Among efficient methods without iterative refinement, our method achieves the best overall results on nearly all benchmarks. Specifically, it attains an $\mathcal{J\&F}$ score of 83.9 on DAVIS2016-M and 84.7 on DAVIS2016, and surpasses the second-best method RCF-Stage1~\cite{RCF} by +6.6 and +6.2 points, respectively.
The substantial improvement highlights the effectiveness of our design in capturing both appearance and motion information for accurate moving object segmentation. 
In comparison with ABR~\cite{xie24appearrefine}, which emphasizes appearance refinement, the proposed method exhibits more robust performance across datasets and metrics, especially in the $\mathcal{F}$ measure, indicating enhanced contour precision and temporal consistency.

It is noteworthy that the proposed model also outperforms several iterative based methods. For instance, compared to OCLR-TTA~\cite{OCLR}, our method achieves a +5.4 point higher $\mathcal{J\&F}$ on DAVIS2016-M. Even when compared with the recent high-capacity SegAnyMotion~\cite{seganymotion}, our method demonstrates competitive results while requiring significantly lower computational cost, which validates the efficiency of the proposed architecture.
On the FBMS-59 dataset, ABR achieves the highest $\mathcal{J}$ score (81.9) due to its appearance-based refinement, yet our method still maintains strong performance (72.5 $\mathcal{J}$, 78.5 $\mathcal{F}$), demonstrating its stable generalization to diverse motion scenarios. 
The performance gap can be attributed to a key difference in approaches: while our model segments based on motion prediction from 4D geometric and camera motion priors, ABR incorporates explicit appearance-based refinement. It gives ABR an advantage in FBMS-59 scenarios where objects have weak motion and mask boundaries are mainly determined by salient appearance. In such case our motion model yields conservative results and leads to incomplete masks. The similar underperformance of SegAnyMotion~\cite{seganymotion} on this benchmark supports this observation. 
Moreover, our model achieves the best results on SegTrackV2, which indicate its robustness to low-resolution and complex motion sequences.

The feed-forward design enables end-to-end inference, strong generalization, and reduced inference cost compared to iterative-based methods. A runtime comparison is listed in the last column of Table~\ref{tab:mos}. Our method maintains high efficient performance at 0.31 seconds per frame, closely matching baselines such as ABR (0.28s) and OCLR-flow (0.20s). In contrast, the iterative-based methods SegAnyMotion and RoMo performs 6.44s and 8.34s per frame, respectively, highlighting the efficiency advantage of our method. The trade-off between accuracy and speed demonstrates the effectiveness of our design in balancing computation with segmentation precision.

Overall, the comparison results verify that the proposed method achieves an excellent trade-off between segmentation quality and computational efficiency. The consistent improvements across multiple datasets and metrics confirm the model’s capability to generalize effectively to various real-world motion segmentation scenarios.

\begin{table}[t]
    \centering
    \caption{
    Comparison with 3D/4D reconstruction-based methods. The best and second-best results are labeled in \textbf{bold} and \underline{underlined}, respectively.
    }
    \scriptsize
    \renewcommand{\tabcolsep}{2.5pt}
    \renewcommand{\arraystretch}{1.1}
    \resizebox{0.5\textwidth}{!}{
    \begin{tabular}{lc|cc|cc|cc} 
    \toprule
    \multirow{2}{*}{Method} & \multirow{2}{*}{Flow} & \multicolumn{2}{c|}{DAVIS2016} & \multicolumn{2}{c|}{DAVIS2017} & \multicolumn{2}{c}{DAVIS-All} \\ 
    \cmidrule(lr){3-4} \cmidrule(lr){5-6} \cmidrule(lr){7-8} 
    & & $\mathcal{J}_M \uparrow$ & $\mathcal{J}_R \uparrow$ & $\mathcal{J}_M \uparrow$ & $\mathcal{J}_R \uparrow$ & $\mathcal{J}_M \uparrow$ & $\mathcal{J}_R \uparrow$\\ 
    \midrule
    DUSt3R~\cite{dust3r} & \cmark & 58.5 & 63.4 & 48.7 & 50.2 & 47.6 & 48.7 \\
    MonST3R~\cite{monst3r} & \cmark & 64.3 & 73.3 & 56.4 & 59.6 & 51.9 & 54.1 \\
    DAS3R~\cite{das3r} & \xmark & 54.2 & 55.8 & 57.4 & 61.3 & 53.9 & 54.8 \\
    Easi3R$_\text{dust3r}$~\cite{chen2025easi3r} & \xmark & 67.9 & 71.4 & 60.1 & 65.3 & 54.7 & 60.6 \\ 
    Easi3R$_\text{monst3r}$~\cite{chen2025easi3r} & \xmark & \underline{70.7} & \underline{79.9} & \underline{67.9} & \underline{76.1} & \underline{63.1} & \underline{72.6} \\
    \textbf{Ours} & \checkmark & \textbf{84.5} & \textbf{96.1} & \textbf{81.1} & \textbf{92.6} & \textbf{74.8} & \textbf{84.5} \\
    \bottomrule
    \end{tabular}
    }
    \label{tab:recon_seg}
\end{table}

\subsubsection{Qualitative Comparison}

Figure~\ref{fig:vis} presents qualitative comparisons by showing multiple visual results across the DAVIS2016-M, DAVIS2017, and SegTrackV2~\cite{segtrackv2} datasets.
Compared with existing methods including OCLR-Flow \cite{OCLR}, SegAnyMotion~\cite{seganymotion} and RoMo~\cite{romo}, the proposed method produces motion masks that are both geometrically complete and visually consistent. It can be found that OCLR-Flow often suffers from fragmented regions and inaccurate boundaries due to its reliance on low-level optical flow cues, while SegAnyMotion exhibits over-segmentation in cluttered scenes, particularly when strong background motion or camera shake is present. RoMo tends to lose fine structural details under large object displacements or illumination variations (e.g., the swan and the ship).
In contrast, our method accurately preserves the object geometry even in challenging conditions such as occlusions, fast motion and background clustering. 
Overall, these visual advantages confirm the robustness and reliability of our proposed design across a wide range of dynamic scenes.

\subsection{Comparison with Reconstruction methods}
We further compare the proposed methods with 3D or 4D scene reconstruction methods that produce explicit motion estimation, including DUSt3R~\cite{dust3r}, MonST3R~\cite{monst3r}, DAS3R~\cite{das3r}, and Easi3R~\cite{chen2025easi3r}. 
We follow Easi3R~\cite{chen2025easi3r} to evaluate our model on the DAVIS datasets~\cite{davis_2016,davis_2017} by using the same evaluation protocols including the mean IoU ($\mathcal{J_M}$) and IoU recall ($\mathcal{J_R}$). All the results of these reconstruction methods follow the reports in Easi3R~\cite{chen2025easi3r}, where all outputs are refined by SAM2~\cite{sam2}.  
As shown in Table~\ref{tab:recon_seg}, the proposed method significantly outperforms all reconstruction-based approaches by a large margin. 
Compared to the second-best method, $Easi3R_\text{monst3r}$, our model improves $\mathcal{J}_M$ by +13.8, +16.2, and +11.7 points, which demonstrates significant superiority in both mask accuracy and region consistency. While the reconstruction works such as DUSt3R and MonST3R rely heavily on explicit flow estimation, the results highlight the effectiveness of our motion-aware learning based architecture design in implicitly disentangling dynamic object motion from camera motion.

\subsection{Ablation Study}

\subsubsection{Effect of Feature Aggregation}
We conduct ablation studies on DAVIS2017 to evaluate the impact of different feature modalities in our feature aggregation module.
The baseline model uses the output of the last two layer in VGB as the fused feature. 
Then additional modalities are incrementally added during training to obtain other models: camera pose features (\textbf{+Cam}), optical flow (\textbf{+Flow}), and shallow-layer features (\textbf{+Shallow}). 
As shown in Table~\ref{tab:ablation}, 
each of them contributes positively to performance. 
The addition of camera features brings a gain of +6.3 points, confirming their importance in suppressing global background motion. Shallow-layer features improve results by +4.5 points, confirming the contribution of object-level structure and appearance to identify coherent regions even when motion signals are weak or ambiguous. The optical flow feature contributes +6.8 points, offering dense local pixel-level motion that complement the geometric representations.
The combinations of modalities further demonstrate their complementary roles. 
Integrating all three modalities yields the highest overall score of 81.4 $\mathcal{J}\&\mathcal{F}$, which validate the effectiveness of our feature fusion strategy.
The results demonstrate that integrating scene geometry, camera pose, and local pixel-level motion into a unified spatio-temporal representation enables the model to segment dynamic regions accurately in a feed-forward manner, without relying on explicit correspondence matching or iterative refinement.

\begin{table}[t]
    \centering
    \caption{
Ablation study on feature aggregation.
Adding the camera pose, optical flow, and shallow-layer features progressively enhances performance on DAVIS2017. The full model, which combines all three modalities, achieves the best overall accuracy, validating the effectiveness of spatio-temporal feature fusion.
    }
    \small
    \setlength{\tabcolsep}{3pt}
    \renewcommand{\arraystretch}{1.1}
    \begin{tabular}{@{}l|ccc|ccc@{}}
    \toprule
    \multirow{2}{*}{Methods} & \multirow{2}{*}{Cam} & \multirow{2}{*}{Flow} & \multirow{2}{*}{Sh.} 
    & \multicolumn{3}{c}{DAVIS2017 } \\
    \cmidrule(l){5-7}
    & & & & $\mathcal{J}\!\&\!\mathcal{F}\uparrow$ & $\mathcal{J}\uparrow$ & $\mathcal{F}\uparrow$ \\
    \midrule
    Baseline        & \xmark & \xmark & \xmark & 67.9 & 68.3 & 67.6 \\
    + Cam            & \cmark & \xmark & \xmark & 74.2 & 74.2 & 74.3 \\
    + Flow           & \xmark & \cmark & \xmark & 74.7 & 74.8 & 74.7 \\
    + Shallow        & \xmark & \xmark & \cmark & 72.4 & 71.9 & 72.9 \\
    + Cam + Shallow    & \cmark & \xmark & \cmark & 76.0 & 76.1 & 75.9 \\
    + Flow + Shallow   & \xmark & \cmark & \cmark & 78.6 & 78.2 & 79.0 \\
    + Cam + Flow       & \cmark & \cmark & \xmark & 80.2 & 80.2 & 80.2 \\
    \textbf{All}    & \cmark & \cmark & \cmark & \textbf{81.4} & \textbf{81.1} & \textbf{81.8} \\
    \bottomrule
    \end{tabular}
    \label{tab:ablation}
\end{table}

\subsubsection{Effect of Data Scale}
We further analyze the impact of data scale on model performance. As shown in Table~\ref{tab:transfer_datasets}, models are progressively trained with +HOI4D~\cite{hoi4d}, +Dynamic Replica~\cite{DynamicReplica}, +OmniWorld-motion, +YTVOS18-m~\cite{xie24appearrefine} and +GOT-Motion. 
As shown in Table \ref{tab:transfer_datasets}, incremental incorporation of the training data consistently enhances generalization across both DAVIS2016-M and DAVIS2017 benchmarks. Starting from HOI4D, adding the data in Dynamic Replica yields a substantial boost of +7.2 / +7.4 ($\mathcal{J}$/ $\mathcal{F}$) on DAVIS2016-M and +8.8 / +8.1 on DAVIS2017, demonstrating the benefit of dynamic scene diversity. By incrementally adding datasets of OmniWorld-motion, YTVOS18-m and GOT-Motion, the performance has been further improved. The combined training over all five datasets achieves 83.5 $\mathcal{J}$ / 84.3 $\mathcal{F}$ on DAVIS2016-M and 81.1 / 81.8 on DAVIS2017. The results demonstrate that our learning architecture for motion segmentation can benefit from larger and more varied, which indicate strong scalability of the proposed framework.

\begin{table}[t]
    \centering
    \caption{
    Ablation study on dataset scale.
Training on progressively larger and more diverse datasets consistently improves segmentation performance, demonstrating the strong scalability and generalization ability of the proposed method.
    }
    \small
    \setlength{\tabcolsep}{3.5pt}
    \renewcommand{\arraystretch}{1.1}
    \resizebox{0.5\textwidth}{!}{
    \begin{tabular}{@{}lccccc@{}}
    \toprule
    \multirow{2}{*}{Training Datasets} & \multirow{2}{*}{\#D.} &
    \multicolumn{2}{c}{DAVIS2016-M} & \multicolumn{2}{c}{DAVIS2017} \\ 
    \cmidrule(lr){3-4} \cmidrule(lr){5-6}
    &  & $\mathcal{J}\uparrow$ & $\mathcal{F}\uparrow$ 
    & $\mathcal{J}\uparrow$ & $\mathcal{F}\uparrow$ \\ 
    \midrule
    + HOI4D ~\cite{hoi4d} & 1 & 67.9 & 68.8 & 65.4 & 66.5 \\
    + Dynamic Replica~\cite{DynamicReplica} & 2 & 75.1 & 76.2 & 73.8 & 74.3 \\
    + OmniWorld-motion & 3 & 77.8 & 77.6 & 76.1& 75.6 \\
    + YTVOS18-m~\cite{xie24appearrefine} & 4 & 80.6 & 80.8 & 79.3 & 79.8 \\
    + GOT-Motion & 5 & \textbf{83.5} & \textbf{84.3} & \textbf{81.1} & \textbf{81.8} \\
    \bottomrule
    \end{tabular}
    }
    \label{tab:transfer_datasets}
\end{table}

\begin{figure}[t]
    \centering
    \includegraphics[width=0.48\textwidth]{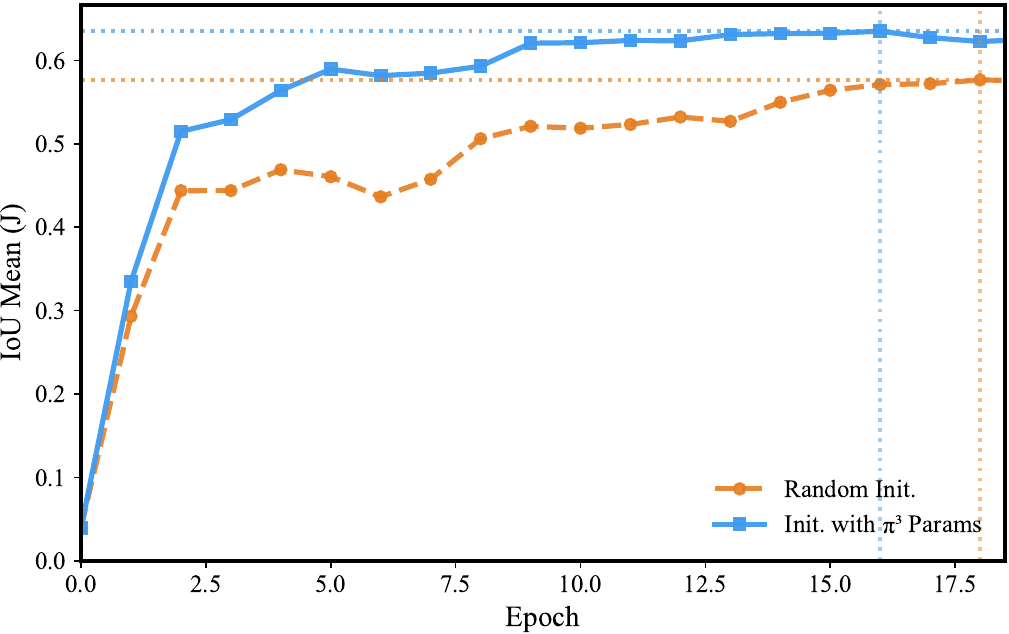}
    \caption{
    Initialization comparison for the motion decoder. Initializing with $\pi^3$ pretrained parameters yields faster convergence and higher IoU compared to random initialization, demonstrating the benefit of large scale geometry pretraining.
    }
    \label{fig:conf_decoder}
    \vspace{-1 em}
\end{figure}

\subsubsection{Effect of Initialization for Motion Decoder}
The confidence decoder in $\pi^3$ is originally trained to predict per-pixel reliability based on reconstruction residuals.
Our training dataset remains small compared to the large-scale data used for pre-training 3D/4D reconstruction models like $\pi^3$. Hence, we leverage the pretrained weights from the confidence decoder in $\pi^3$ to initialize our motion decoder module. The strategy promotes more stable training and faster convergence
Its comparison with the random initialization is shown in Figure~\ref{fig:conf_decoder}. It can be found that the initialization with $\pi^3$ parameters accelerates optimization (reaching a given $\mathcal{J}$ in fewer epochs) and improves the final accuracy. The results suggest that pretraining from large 4D scenes provides a geometrically meaningful parameterization that benefits motion estimation.

\section{Conclusion}

In this paper, we present GeoMotion, a geometry-informed feed-forward framework for motion segmentation that unifies 2D pixel-level motion and 4D geometric priors within an end-to-end architecture. By leveraging latent geometry representations from pretrained reconstruction models such as $\pi^3$, our method learns to implicitly disentangle object and camera motion without relying on explicit correspondence estimation or iterative optimization. The proposed feature aggregation module and motion decoder integrates 4D geometric, optical, and camera pose information to achieve precise and temporally consistent motion segmentation. Extensive experiments across multiple benchmarks demonstrate that the proposed GeoMotion achieves state-of-the-art performance while maintaining high efficiency. The work opens new directions toward fully feed-forward, geometry-guided motion perception and dynamic 4D scene understanding, bridging the gap between reconstruction and motion segmentation.
{
    \small
    \bibliographystyle{ieeenat_fullname}
    \bibliography{main}
}

\clearpage
\maketitlesupplementary
\appendix

\section{Architecture Details}
Algorithm~\ref{alg:motion_aware_decoder} presents the detailed pseudo-code of the proposed Feature Aggregation Module, providing implementation-level clarification of the fusion and projection steps.

\begin{algorithm}[t]
\caption{Feature Aggregation Module}
\label{alg:motion_aware_decoder}
\KwIn{
$F_{\text{geo}}^{\text{low}}$: low-level geometry features $[B\!*\!N,\; hw,\; 2C]$ \\
$F_{\text{geo}}^{\text{high}}$: high-level geometry features $[B\!*\!N,\; hw,\; 2C]$ \\
$F_{\text{cam}}$: camera features $[B\!*\!N,\; hw,\; 512]$ \\
$\text{flow}$: optical flow map $[B,\;N,\;H,\;W]$ \\
$C$: 1024
}
\KwOut{
$F_{\text{fuse}}$: fused features $[B\!*\!N,\; hw,\; 2048]$
}


\BlankLine
\tcp{1) Concatenate multi-level geometry features}
$F_{\text{geo}} \gets \texttt{Concat}(F_{\text{geo}}^{\text{low}}, F_{\text{geo}}^{\text{high}})$ \tcp*{$[B\!*\!N,\; hw,\; 4C]$}

\tcp{2) Project geometry features to 2048-d}
$F_{\text{geo}} \gets \texttt{Linear+ReLU}(F_{\text{geo}})$ \tcp*{$[B\!*\!N,\; hw,\; 2048]$}

\BlankLine
\tcp{3) Encode optical flow and convert to patch tokens}
$F_{\text{flow}} \gets 
    \texttt{BilinearDown}(
        \texttt{CNN}(\text{flow}),
        h,\;w
    )
$ \tcp*{$[B\!*\!N,\; hw,\; 128]$}

\BlankLine
\tcp{4) Fuse geometry, flow, and camera features}
$F_{\text{cat}} \gets \texttt{Concat}(F_{\text{geo}}, F_{\text{flow}}, F_{\text{cam}})$ \tcp*{$[B\!*\!N,\; hw,\; 2688]$}

\tcp{5) Final projection}
$F_{\text{fuse}} \gets \texttt{Linear}(F_{\text{cat}})$ \tcp*{$[B\!*\!N,\; hw,\; 2048]$}

\Return $F_{\text{fuse}}$
\end{algorithm}
\vspace{-0.4em}

\section{Ablation for SAM2}
\begin{figure}[t]
    \centering
    \includegraphics[width=0.95\linewidth]{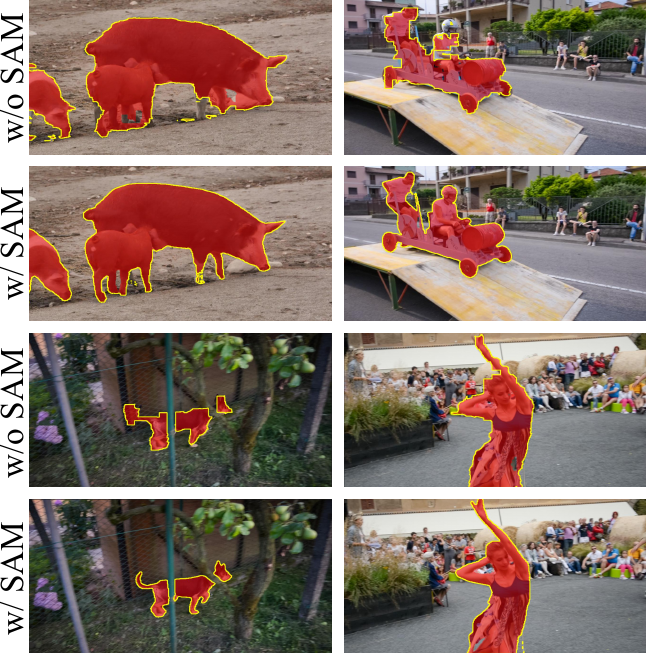}
    \caption{Qualitative ablation results for SAM2.}
    \label{fig:sam_qual}
    \vspace{-1em}
\end{figure}

The results of SAM2 ablation are shown in Tab.~\ref{tab:sam_ablation} and Fig.~\ref{fig:sam_qual}. SAM2 mainly enhances boundary quality, and our core method remains effective without it. The raw output of the model still surpasses other refined methods like Easi3R w/SAM2.

\begin{table}[t]
\centering
\scriptsize
\setlength{\tabcolsep}{1.6pt}
\renewcommand{\arraystretch}{1.0}
\scalebox{1}{%
\begin{tabular}{l|cc|cc|cc}
\toprule
\textbf{Method} &
\multicolumn{2}{c|}{JM$\uparrow$} &
\multicolumn{2}{c|}{JR$\uparrow$} &
\multicolumn{2}{c}{FM$\uparrow$} \\
& w/o SAM & w/ SAM & w/o SAM & w/ SAM & w/o SAM & w/ SAM \\
\midrule
Easi3R$_{\text{dust3r}}$~\cite{chen2025easi3r}  & 46.86 & 60.1 & 50.54 & 65.3 & 39.06 & -- \\
Easi3R$_{\text{monst3r}}$~\cite{chen2025easi3r}  & 54.75 & 67.9 & 66.16 & 76.1 & 44.09 & -- \\
MonST3R~\cite{monst3r}                   & 38.07 & 56.4 & 36.05 & 59.6 & 48.24 & -- \\
DAS3R~\cite{das3r}                      & 44.51 & 57.4 & 43.95 & 61.3 & 46.71 & -- \\
VGGT4D~\cite{vggt4d} & 56.45 & --   & 65.62 & --   & 51.09 & -- \\
\midrule
OCLR-flow~\cite{OCLR}                 & 69.90 & --   & --    & --   & --    & -- \\
ABR~\cite{xie24appearrefine}          & 74.60 & --   & --    & --   & --    & -- \\
\midrule
Ours                      & \textbf{75.38} & \textbf{81.13} & \textbf{87.19} & \textbf{92.63} &  72.29 & 81.82 \\
\bottomrule
\end{tabular}}
\caption{Ablation for SAM2 on DAVIS-2017.}
\label{tab:sam_ablation}
\end{table}

\section{Qualitative Comparison with Reconstruction Methods}

\begin{figure*}[h]
    \centering
    \includegraphics[width=0.93\linewidth]{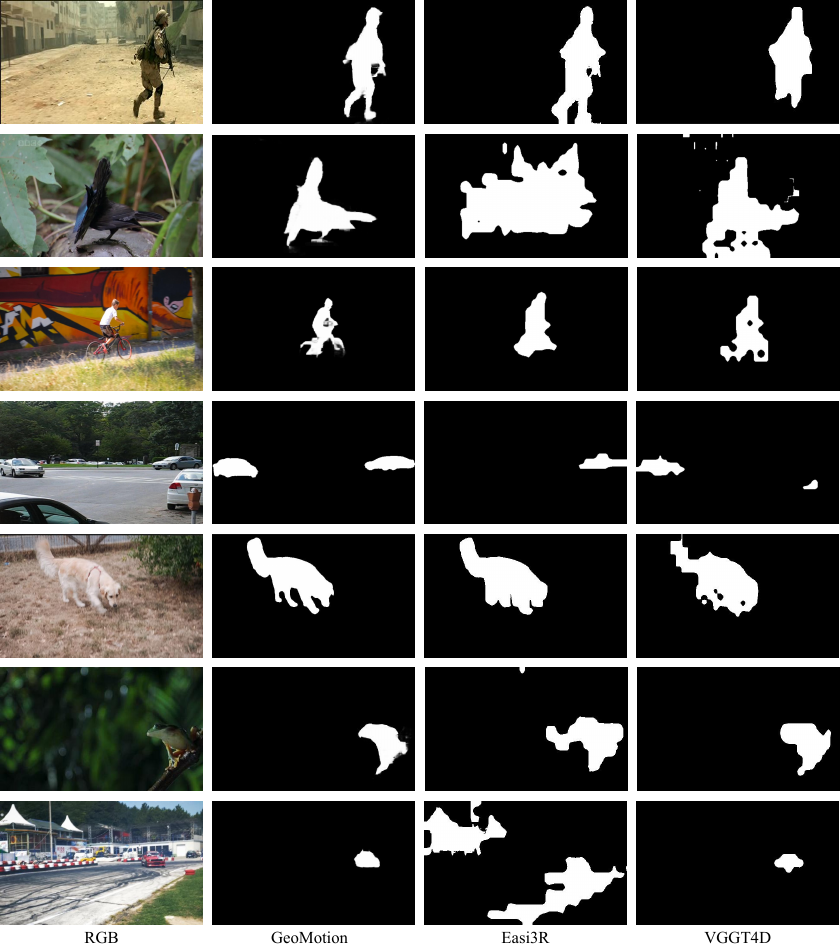}
    \caption{Compare with Easi3R and VGGT4D without SAM2.}
    \label{fig:vggt4d}
    \vspace{-1 em}
\end{figure*}

Figure~\ref{fig:vggt4d} shows qualitative comparisons with reconstruction-based methods, including Easi3R~\cite{chen2025easi3r} and VGGT4D~\cite{vggt4d}, using raw predictions without SAM2 refinement. 
GeoMotion produces cleaner and more compact motion masks with fewer background false positives. In contrast, reconstruction-based approaches may over-segment background regions or generate fragmented masks under weak motion cues. Our geometry-, flow-, and camera-aware fusion results in more stable and coherent object-level predictions.

\section{More Visualizations on Dynamic Scenes}
\begin{figure*}[t]
    \centering  \includegraphics[width=0.9\textwidth]{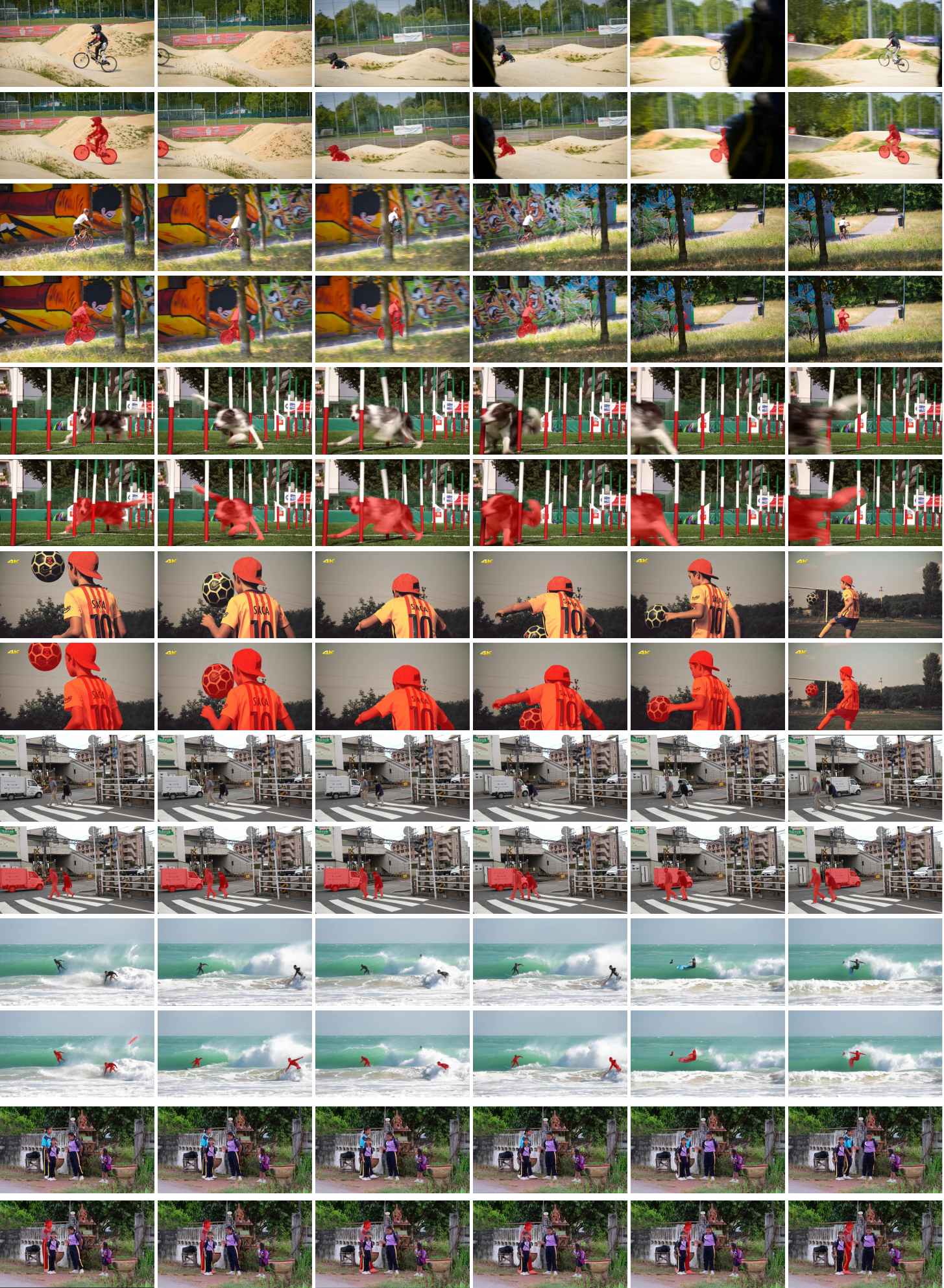}
    \caption{
    \textbf{More visual examples of dynamic masks predicted by GeoMotion on the DAVIS benchmark.} 
    Odd rows show the RGB input frames, while even rows present the corresponding predicted dynamic masks.
    }
    \label{fig:more_vis}
\end{figure*}

We provide additional qualitative visualizations in Fig.~\ref{fig:more_vis} to further illustrate the robustness of GeoMotion.
Seven representative sequences are selected from the DAVIS benchmark: the first three depict scenes with a single moving object, while the remaining four contain multiple moving objects.
Across diverse challenging scenarios such as motion blur, heavy occlusion, fast object motion, large camera movement, and strong appearance similarity, GeoMotion consistently produces accurate and temporally stable dynamic masks. These results indicate that the geometry-driven representation effectively captures motion cues even when conventional signals are unreliable.

\end{document}